\documentclass[conference,a4paper]{IEEEtran}
\ifdefined\XeTeXversion
  \usepackage[T1]{fontenc}
\fi
\usepackage[cmex10]{amsmath}
\usepackage{amssymb}
\usepackage{url}
\usepackage{graphicx}
\usepackage{cite}
\usepackage{pgfplots}
\usepackage{pgfplotstable}
\usepgfplotslibrary{groupplots}
\pgfplotsset{compat=1.18}

\begin{document}
\bstctlcite{IEEEexample:BSTcontrol}

\title{\uppercase{Static Land-Cover and Fire-Related Information in Low-Dimensional TESSERA Embeddings}}

\author{
    \IEEEauthorblockN{Jing Zhang}
    \IEEEauthorblockA{
        \textit{Lancaster University College at Beijing Jiaotong University}\\
        Weihai, China
    }
    \and
    \IEEEauthorblockN{Zhengpeng Feng\textsuperscript{*}}
    \IEEEauthorblockA{
        \textit{Department of Computer Science and Technology}\\
        University of Cambridge\\
        Cambridge, United Kingdom\\
        \textit{*Corresponding author: zf281@cam.ac.uk}
    }
}

\maketitle

\begin{abstract}
We compare two low-dimensional projections of annual Tessera embeddings for
land-cover classification and for distinguishing burned locations from matched
controls. Using Tessera v1.1 embeddings from 2018 to 2021, we fit a stable
principal component basis to four-year location means and a dynamic basis to
annual residuals. In the land-cover experiment, both bases use the same
training locations. At 16 dimensions, mean macro-F1 across five spatial
splits is 0.6693 for the stable projection and 0.5713 for the dynamic
projection. The fire experiment uses 4,351 matched fire-control pairs from
Monitoring Trends in Burn Severity (MTBS), and here too both bases are fitted
to the same training fire and control locations. At 16 dimensions, the
dynamic projection leads by 0.0215 in area under the receiver operating
characteristic curve (AUROC) for change magnitude and by 0.0180 in macro-F1
for signed differences. Both metrics give equal weight to each held-out
component. These gaps narrow at 32 dimensions and are below 0.001 on both
measures at 64.
\end{abstract}

\begin{IEEEkeywords}
Satellite embeddings, dimensionality reduction, land-cover classification,
fire-related change detection.
\end{IEEEkeywords}

\section{Introduction}
\label{sec:intro}

Several satellite foundation models are now distributed as precomputed
embeddings, so that a user can map a region without handling raw imagery
\cite{feng2026tessera,brown2025alphaearth}. Tessera encodes a year of Sentinel-1 and
Sentinel-2 observations at each 10\,m pixel into a 128-dimensional embedding
\cite{feng2026tessera,feng2026applications}, building on a pixel-wise Barlow
Twins model for cloud-corrupted time series \cite{lisaius2024barlow}.
AlphaEarth Foundations provides similar annual embeddings with 64 dimensions
\cite{brown2025alphaearth}. These embeddings have been used to map tree
species \cite{ball2026species}, crop types \cite{lisaius2026senegal}, burned
area \cite{knezevic2026wildfire}, and other environmental variables
\cite{feng2026applications}.

Although compact relative to the imagery they summarize, these embeddings are
still large: one year of 128-dimensional int8 embeddings for the conterminous
United States amounts to about 10\,TB. Tessera v2 reduces this cost with
Matryoshka embeddings, whose 16-dimensional prefix keeps about 92\% of the
full 128-dimensional performance \cite{feng2026tesserav2}. Earlier releases,
including the Tessera v1.1 embeddings that we use \cite{feng2026hpc}, have no
such ordering \cite{feng2026tesserav2}, so a user must choose a projection.
Principal component analysis (PCA) is a natural choice because it needs no
labels.

A PCA basis keeps the directions of greatest variance in the data to which it
is fitted, so the fitting data determine what a projection retains. Annual
embeddings vary in two ways: persistently between locations, for instance in
land cover, and from year to year at one location, through disturbance,
acquisition effects, and other interannual differences. Year-to-year
variation affects downstream use. A tree species classifier trained on one
year of Tessera v1 or AlphaEarth embeddings loses accuracy in another year
\cite{ball2026species}, and Tessera v1.1 was retrained for greater stability
across years \cite{feng2026hpc}. One might expect persistent directions to
serve land-cover classification and year-to-year directions to serve
disturbance detection. This is not guaranteed, because a fire that turns
forest into shrubland can also move the embedding along land-cover directions,
and the leading year-to-year directions may mainly reflect acquisition
effects. To our knowledge, no prior work has tested which kind of direction a
low-dimensional projection should keep for each task.

Projections of this kind have a long history in change detection. When PCA is
applied to multi-date imagery, the leading components mainly describe the
unchanged surface and minor components describe change
\cite{richards1984pca,fung1987pca}. Continuous change detection and
classification (CCDC) instead models temporal observations directly
\cite{zhu2014ccdc}, and foundation model features have been adapted to
bitemporal change detection \cite{chen2024ttp}. More recent work shows that
annual Tessera embeddings encode wildfire disturbance
\cite{knezevic2026wildfire} and that forecast embeddings can flag anomalous
surface change without labels \cite{li2026forecasting}. We carry the
multitemporal PCA idea over to embeddings.

Specifically, we split each annual embedding into a four-year location mean
and an annual residual, and fit one PCA basis to each, giving a stable and a
dynamic projection. We compare the two at equal dimension, with random
projections and the full embedding as references, on two tasks: static
land-cover classification against Land Change Monitoring, Assessment, and
Projection (LCMAP) reference labels \cite{lcmap_reference}, and
discrimination of burned locations mapped by Monitoring Trends in Burn
Severity (MTBS) \cite{mtbs_overview} from matched controls. At 16 dimensions,
the stable projection gives the higher land-cover macro-F1 (0.6693 versus
0.5713). On the fire task, the dynamic projection gives the higher macro-F1
(0.8445 versus 0.8265) and the higher area under the receiver operating
characteristic curve, AUROC (0.8133 versus 0.7917), when both bases are
fitted to the same fire and control training sites. In both tasks, the gap
narrows as more dimensions are retained.

\section{Methods}
\label{sec:methods}

\subsection{Embeddings and PCA bases}
\label{sec:bases}

We use annual 128-dimensional Tessera v1.1 embeddings
\cite{feng2026tessera,feng2026hpc} from the dClimate Zarr archive on Source
Cooperative, and recover each vector by multiplying its stored 8-bit signed
integers by the pixel-year scale. For the embedding
$\mathbf e_{i,t}\in\mathbb R^{128}$ at location $i$ in year $t$, we define the
location mean and the annual residual as
\begin{equation}
\begin{aligned}
\boldsymbol\mu_i
&= \frac{1}{4}\sum_{t\in\mathcal T}\mathbf e_{i,t},\\
\boldsymbol\delta_{i,t}
&= \mathbf e_{i,t}-\boldsymbol\mu_i,\\
\mathcal T
&= \{2018,2019,2020,2021\}.
\end{aligned}
\label{eq:decomposition}
\end{equation}
Intuitively, the mean records what persists at a location and the residual
records how each year departs from it. The 2017 embeddings serve only as the
previous year for 2018 fire events.

\begin{figure*}[!t]
\centering
\includegraphics[width=\textwidth]
{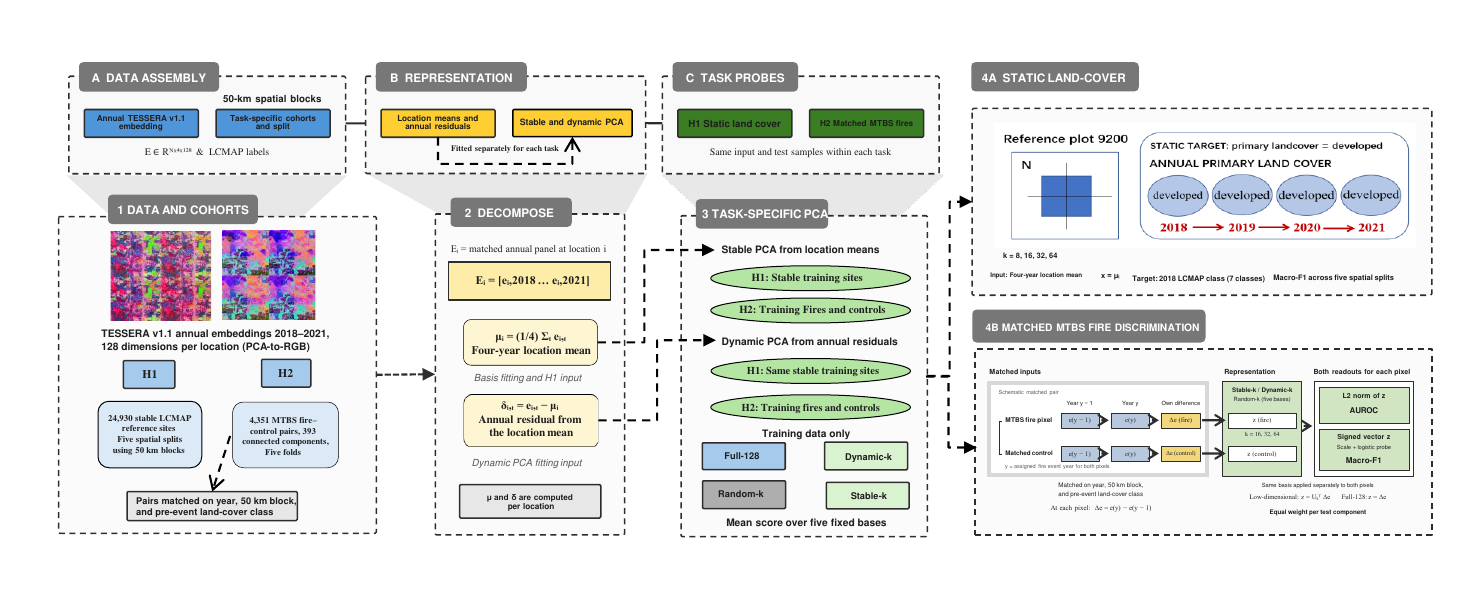}
\caption{Each annual embedding is split into a location mean and annual
residuals, and each part receives its own PCA basis. In H1, the mean
$\boldsymbol\mu$ is projected to $\mathbf z$. H2 pairs share a 50-km block
and pre-event land-cover class, with separation $\leq 50$ km. The 2017
embeddings supply the prior-year endpoint only for 2018 events.}
\label{fig:lcmap-workflow}
\end{figure*}

PCA finds orthogonal directions that successively retain the greatest
variance in its fitting data \cite{jolliffe2016pca}. We fit it separately to
centred location means and to annual residuals, which gives the stable and the
dynamic basis. The leading $k$ directions of basis $b\in\{s,d\}$ form
$U_b^{(k)}\in\mathbb R^{128\times k}$. PCA uses centred inputs without
per-channel scaling or per-vector $\ell_2$ normalization. Each task fits its
own bases on its training partition only, so validation and test locations
never contribute to PCA fitting. Both tasks require finite embeddings in every
required year, and the fire cohort also excludes all-zero vectors. We call the
two tasks H1 (static land-cover classification) and H2 (matched fire-control
discrimination), and Fig.~\ref{fig:lcmap-workflow} summarizes both.

\subsection{Static land-cover classification (H1)}
\label{sec:static-methods}

H1 uses 24,930 locations in the conterminous United States (CONUS) whose LCMAP
reference labels remain stable during 2018--2021 \cite{lcmap_reference}. These
are interpreted reference labels, distinct from the raster land-cover product
used for matching in H2. The target is the 2018 class among Barren, Cropland,
Developed, Grass/Shrub, Tree Cover, Water, and Wetland, and we exclude
locations labelled Snow/Ice in any study year.

Both bases are fitted at the same training locations, with stable PCA using
their four-year means and dynamic PCA using their annual residuals. Whichever
basis is used, the classifier receives the projected mean
\begin{equation}
\mathbf z_{i,b}^{(k)}
=(U_b^{(k)})^\top(\boldsymbol\mu_i-\mathbf c_b),
\label{eq:static-projection}
\end{equation}
where $\mathbf c_b$ is the centre estimated when fitting basis $b$. The input
summary, target labels, and test locations are therefore the same for both
bases at a given dimension.

We evaluate $k\in\{8,16,32,64\}$ over five spatial splits. In each split, a
seeded hash assigns whole 50-km blocks (EPSG:5070) to training, validation,
and testing in target proportions of 60\%, 20\%, and 20\%, so all years at a
location fall in one partition. Grouping by block accounts for known spatial
dependence in the evaluation \cite{roberts2017crossvalidation}. A
class-balanced multinomial logistic classifier predicts land cover from each
representation.

\subsection{Matched fire-control discrimination (H2)}
\label{sec:mtbs-methods}

H2 uses 4,351 matched pairs from 947 MTBS incidents in the CONUS study area.
Fire locations have moderate or high mapped burn severity, as defined in the
MTBS product descriptions \cite{mtbs_products}, and belong to incidents of at
least 1,000 acres. This size threshold is our own restriction, not an MTBS rule. MTBS
uses different minimum mapping sizes in the eastern and western United States
\cite{mtbs_overview}.

Controls come from the stable LCMAP reference pool of H1 and take the event
year of their paired fire location. A control must share the fire location's
50-km block, lie within 50 km of it, and have the same class in the year
before the event in LCMAP Collection 1.3 primary land cover
\cite{usgs2022lcmap13}. These conditions use event-source and reference-point
coordinates, and an independent check after matching confirmed the pre-event
class agreement at the Tessera pixel centres of all 4,351 pairs. Matching
first maximizes the number of pairs and then minimizes their total distance,
using each control at most once. Controls must also keep unchanged mapped land
cover throughout 2018--2021 and across their adjacent-year comparison, and lie
outside all MTBS perimeters mapped in 2018--2021 and their 1-km buffers. These
checks exclude mapped fires and land-cover transitions in the specified
records, but they do not establish the absence of every disturbance.

Pairs that share an incident, a 50-km block, a location ID, or a native pixel
are linked, which yields 393 connected components. We assign intact
components, largest first, to whichever of five bins holds the fewest pairs,
breaking ties between equal sizes with a fixed seed. Each fold tests one bin,
validates on the next, and trains on the other three, so every component is
tested once. In each fold, both bases are fitted to the same training fire and
control sites. Each site contributes one four-year mean to stable PCA and four
annual residuals to dynamic PCA. We adopted this same-site design after an
initial analysis that fitted the stable basis to control means only. We
therefore refitted the stable basis and its signed-difference classifier with
the procedure of Section~\ref{sec:model-selection}, and kept the held-out
predictions of the dynamic basis and of the random and full-dimensional
references.

For the assigned event year $y$, each fire and control location supplies the
same type of adjacent-year difference and its projection:
\begin{equation}
\begin{aligned}
\Delta\mathbf e_{i,y}
&= \mathbf e_{i,y}-\mathbf e_{i,y-1},\\
\mathbf z_{i,y,b}^{(k)}
&= (U_b^{(k)})^\top\Delta\mathbf e_{i,y}.
\end{aligned}
\label{eq:jump}
\end{equation}
We evaluate two readouts of this projection. Its norm
$\|\mathbf z_{i,y,b}^{(k)}\|_2$ measures change magnitude and serves as the
ranking score for AUROC. A class-balanced binary logistic classifier uses the
signed vector to predict fire versus control and is evaluated by macro-F1, so
this readout keeps the individual component values rather than one magnitude.
Both readouts are primary at $k=16$, and $k=32$ and $k=64$ are secondary
comparisons. Across the five folds, every representation and dimension is
scored on the same 8,702 held-out samples from the 393 components.

\subsection{References, model selection, and uncertainty}
\label{sec:model-selection}

At each dimension, we compare both PCA bases with five prespecified orthonormal
random bases, obtained by QR decomposition of Gaussian matrices and fixed
across evaluations, with nested prefixes for the tested dimensions. The random
reference is the mean of their metric scores, not a prediction ensemble or the
best-performing seed. The full-dimensional reference, Full-128, uses the
unprojected location mean for H1 and the unprojected difference vector or its
norm for the two H2 readouts.

Logistic regression provides a common linear readout \cite{hastie2009esl}. We
standardize its inputs with training means and population standard deviations,
and fit class-balanced, $\ell_2$-regularized models with L-BFGS. Validation
macro-F1 selects $C$ from $\{0.01,0.1,1,10\}$ for H1 and from
$\{0.01,0.1,1,10,100\}$ for H2. For each H2 value of $C$, we also choose the
threshold that maximizes pooled validation macro-F1, searching the midpoints
between consecutive distinct validation probabilities and values just outside
their range. Ties favour the smallest threshold within each $C$ and the
smallest $C$ across values. The selected classifier remains fitted to the
training partition, and the test set is used only for evaluation.

H1 macro-F1 is computed over each complete seven-class test set and summarized
by its mean and standard deviation across the five splits. On the first split,
which we designate the primary split, 1,000 paired bootstrap draws of spatial
blocks recompute the global metric and the stable-minus-dynamic difference.
This interval describes the primary split, not uncertainty in the five-split
mean. For H2, each readout is computed within a held-out component and then
averaged with equal weight across components, so components with more sampled
locations do not automatically dominate. Paired differences use 1,000
bootstrap resamples of components. Both procedures follow the general
bootstrap framework \cite{efron1979bootstrap}, with resampling units chosen to
match our evaluation design, and both resample fixed test outputs without
refitting. We take the 2.5th and 97.5th percentiles of the paired differences
as nominal 95\% intervals. These intervals have no multiplicity adjustment,
and we treat secondary comparisons as descriptive.

\section{Results and Discussion}
\label{sec:results}

\subsection{Static land-cover classification}

At every tested dimension, the stable projection has higher mean macro-F1 than
both the dynamic projection and the five-basis random mean
(Table~\ref{tab:static}). At $k=16$, its mean macro-F1 across five spatial
splits is 0.6693, compared with 0.5713 for the dynamic projection and 0.5886
for the random reference, with standard deviations across splits of 0.0107,
0.0071, and 0.0062. On the primary split, the paired stable-minus-dynamic
difference is 0.0981, with a nominal 95\% interval of [0.0856, 0.1118].

The gap is largest at the smallest tested dimension. The mean
stable-minus-dynamic gap is 0.1334 at eight dimensions, 0.0408 at 32, and
0.0047 at 64, where the primary-split interval includes zero. Even the best
16-dimensional projection remains below the full representation, whose mean
macro-F1 is 0.7782.

\begin{table}[t]
\centering
\caption{The stable projection leads at every tested dimension. Static
land-cover macro-F1 across five spatial splits, as mean $\pm$ standard
deviation. The random reference averages scores from five bases within each
split.}
\label{tab:static}
\small
\setlength{\tabcolsep}{4pt}
\begin{tabular}{rccc}
\hline
$k$ & Stable & Dynamic & Random mean \\
\hline
8
& $0.5614\pm0.0068$
& $0.4280\pm0.0061$
& $0.4530\pm0.0071$ \\
16
& $0.6693\pm0.0107$
& $0.5713\pm0.0071$
& $0.5886\pm0.0062$ \\
32
& $0.7197\pm0.0123$
& $0.6789\pm0.0125$
& $0.6834\pm0.0124$ \\
64
& $0.7490\pm0.0120$
& $0.7443\pm0.0128$
& $0.7386\pm0.0127$ \\
\hline
128 & \multicolumn{3}{c}
{Full representation: $0.7782\pm0.0154$} \\
\hline
\end{tabular}
\end{table}

\subsection{Matched fire-control discrimination}

At 16 dimensions, the dynamic projection separates fires from controls better
than the stable projection on both readouts (Table~\ref{tab:mtbs}). For change
magnitude, its AUROC is 0.8133 against 0.7917, a paired difference of 0.0215
(nominal 95\% interval [0.0131, 0.0318]). For the signed difference vector,
its macro-F1 is 0.8445 against 0.8265, a difference of 0.0180
[0.0042, 0.0309]. The dynamic projection also exceeds the five-basis random
mean on both readouts.

\begin{figure}[!htb]
\centering
\pgfplotstableread{
k  gap  lower  upper
16 0.021520067467 0.013057557180 0.031837916130
32 0.001939283351 -0.001434094211 0.005520872704
64 0.000779918214 -0.001602945763 0.003765554269
}\MTBSNorm
\pgfplotstableread{
k  gap  lower  upper
16 0.017983192495 0.004222751141 0.030910162727
32 0.008754834250 0.000394724442 0.017439483691
64 0.000465623213 -0.005117548166 0.006051946203
}\MTBSSigned
\definecolor{mtbsrose}{RGB}{179,57,92}
\newcommand{\MTBSfont}{\fontencoding{OT1}\fontfamily{cmss}\fontseries{m}\selectfont\scriptsize}
\newcommand{\MTBSbold}{\fontencoding{OT1}\fontfamily{cmss}\fontseries{bx}\selectfont\scriptsize}
\newcommand{\MTBStrace}[1]{%
  \addplot[gray!85,dashed,line width=0.6pt,mark=none]
    coordinates {(-0.24,0) (2.24,0)};
  \addplot+[mtbsrose,mark=*,mark size=1.9pt,line width=1.05pt,
    mark options={solid,draw=mtbsrose,fill=mtbsrose},
    error bars/.cd,y dir=both,y explicit,
    error bar style={line width=0.85pt,solid,draw=mtbsrose},error mark=none]
    table[x expr=\coordindex,y=gap,
      y error plus expr=\thisrow{upper}-\thisrow{gap},
      y error minus expr=\thisrow{gap}-\thisrow{lower}] {#1};
  \addplot[only marks,mark=-,mark size=2.3pt,mtbsrose,line width=0.85pt]
    table[x expr=\coordindex,y=lower] {#1};
  \addplot[only marks,mark=-,mark size=2.3pt,mtbsrose,line width=0.85pt]
    table[x expr=\coordindex,y=upper] {#1};
}
\begin{tikzpicture}
\begin{groupplot}[
  group style={group size=2 by 1, horizontal sep=5mm},
  width=0.38\columnwidth,height=0.19\columnwidth,scale only axis,
  axis lines=left,axis line style={-},tick align=outside,
  xmin=-0.24,xmax=2.24,ymin=-0.008,ymax=0.035,
  xtick={0,1,2},xticklabels={16,32,64},
  ytick={0,0.01,0.02,0.03},scaled ticks=false,
  yticklabel style={/pgf/number format/fixed,/pgf/number format/precision=2,
                    /pgf/number format/fixed zerofill},
  ymajorgrids,grid style={black!13,line width=0.35pt},
  ticklabel style={font=\MTBSfont},
  label style={font=\MTBSfont},
  title style={at={(axis description cs:0,1.01)},anchor=south west,
               inner sep=0pt,font=\MTBSbold},
  enlarge x limits=false]
\nextgroupplot[title={(a) Change-norm AUROC}]
\MTBStrace{\MTBSNorm}
\nextgroupplot[title={(b) Signed macro-F1},yticklabels=\empty]
\MTBStrace{\MTBSSigned}
\end{groupplot}
\node[rotate=90,anchor=south,font=\MTBSfont] at ([xshift=-19pt]group c1r1.west)
  {Dynamic $-$ stable};
\path (group c1r1.south west) -- node[midway,below=10pt,font=\MTBSfont]
  {Dimension $k$} (group c2r1.south east);
\end{tikzpicture}
\caption{The advantage of the dynamic projection on MTBS shrinks as dimension
grows. Dynamic-minus-stable differences, with both bases fitted to the same
training fire and control sites. Error bars are nominal 95\% paired
component-bootstrap intervals on fixed held-out predictions. The dashed line
marks zero difference.}
\label{fig:mtbs-differences}
\end{figure}
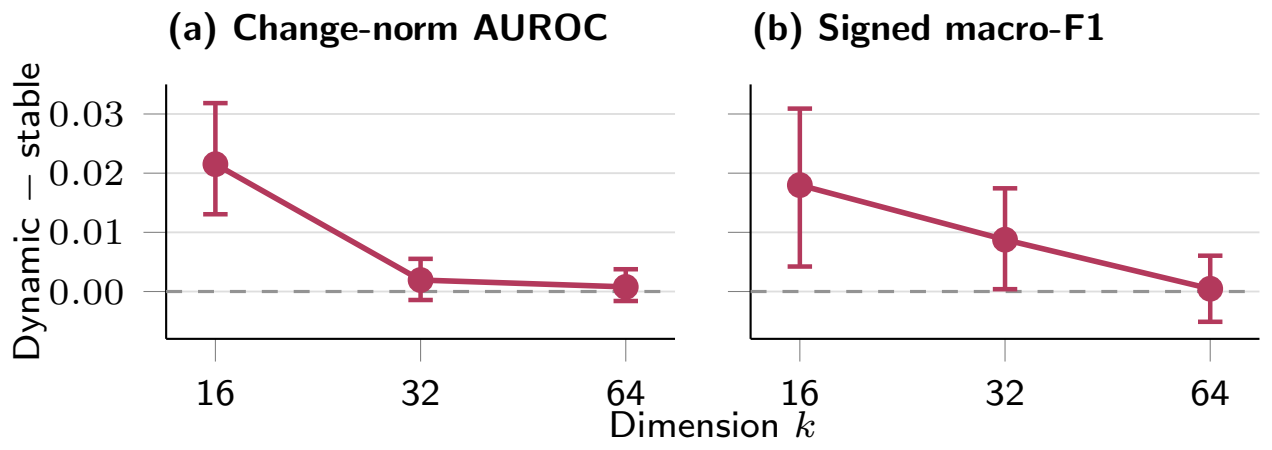

\begin{table}[t]
\centering
\caption{The dynamic projection leads at 16 dimensions, and the bases converge
by 64. MTBS scores averaged with equal weight across the same 393 held-out
components, with both bases fitted to the same training sites. AUROC uses the
norm of the difference vector, and macro-F1 comes from the classifier that
uses the signed vector. Differences are calculated from unrounded scores.}
\label{tab:mtbs}
\small
\setlength{\tabcolsep}{4pt}
\begin{tabular}{lrccc}
\hline
Metric & $k$ & Stable & Dynamic & Random mean \\
\hline
AUROC    & 16 & 0.7917 & 0.8133 & 0.7750 \\
         & 32 & 0.8057 & 0.8076 & 0.7849 \\
         & 64 & 0.8016 & 0.8023 & 0.7943 \\
\hline
Macro-F1 & 16 & 0.8265 & 0.8445 & 0.7537 \\
         & 32 & 0.8650 & 0.8738 & 0.8290 \\
         & 64 & 0.8848 & 0.8853 & 0.8755 \\
\hline
\multicolumn{5}{c}
{Full 128: AUROC 0.8007, macro-F1 0.9060} \\
\hline
\end{tabular}
\end{table}

These differences shrink as more dimensions are retained
(Fig.~\ref{fig:mtbs-differences}). At 32 dimensions, the AUROC gap falls to
0.0019 [$-0.0014$, 0.0055], and the macro-F1 gap is 0.0088 [0.0004, 0.0174],
with the lower bound just above zero. At 64 dimensions, the gaps are 0.0008
for AUROC and 0.0005 for macro-F1, and both intervals include zero. Note that
the full 128-dimensional difference vector gives the highest signed macro-F1
(0.9060), but for change magnitude the 16-dimensional dynamic projection has
a higher AUROC than the full difference norm (0.8133 versus 0.8007).

\subsection{Interpretation and limitations}
\label{sec:limits}

Our results complement recent analyses of Earth embeddings. Knezevic et al.\
show that annual Tessera embeddings encode wildfire disturbance
\cite{knezevic2026wildfire}, Guerrero-Viu et al.\ analyse the temporal
sensitivity of Tessera embeddings \cite{guerrero2026temporal}, and Young and
Keshav \cite{young2026earth} and Benson et al.\
\cite{benson2026decomposing} decompose embeddings into interpretable parts.
Multitemporal PCA has long been used to map fire damage
\cite{richards1984pca}. In contrast, we compare two PCA projections of the
same embeddings at equal dimension. Differences in cohort, burn severity
labels, and evaluation units limit direct comparison with published scores.

Our study has several limitations. First, H1 compares projections at
retrospectively selected stable locations, with training locations and
pre-projection inputs held fixed. Because the bases are fitted separately for
each task, their spans can overlap.

Second, the H2 comparison was not prespecified. We adopted the same-site
design after an initial analysis (Section~\ref{sec:mtbs-methods}), and on the
same MTBS cohort it favours the dynamic projection on both readouts at 16
dimensions. H2 also uses the full 2018--2021 window for decomposition and
control eligibility, so its results describe retrospective discrimination.
Its controls can contain disturbances absent from the MTBS record.

Third, other kinds of change gave mixed results. Exploratory tests with
active-fire detections, settlement expansion, and flood events varied, and led
us to focus on MTBS. In a separate LCMAP analysis that treated land-cover
transitions as positives and included land-use-only transitions among the
negatives, the stable projection had the higher macro-F1 (0.4211 versus
0.4146) and AUROC (0.7505 versus 0.7296) at 16 dimensions. That analysis uses
a different protocol, so the role of change type remains uncertain.

Finally, spatial dependence may persist across neighbouring blocks, because
the splits use no additional distance buffer
\cite{roberts2017crossvalidation}. Reprojecting the archived longitude and
latitude changes seven H1 block assignments at boundaries. We retained the
stored block IDs, which are consistent with the archived projected
coordinates, and found no native pixels shared across partitions. The
bootstrap intervals are conditional on the fitted predictions, and
generalization to other regions and event types remains to be tested.

\section{Conclusion}
\label{sec:conclusion}

On the static land-cover task, the stable projection gives higher mean
macro-F1 than the dynamic projection at every tested dimension. On matched
MTBS fires, the dynamic projection gives higher AUROC and macro-F1 at 16
dimensions when both bases are fitted to the same training sites. Its
advantage is smaller at 32 dimensions, and neither readout clearly favours one
basis at 64. In both tasks, the largest gap occurs at the smallest tested
dimension, so the choice of basis matters most when the projection is most
compact. These findings motivate a comparison between projections fitted after
training and the learned nested prefixes of Tessera v2
\cite{feng2026tesserav2}.

\clearpage
\bibliographystyle{IEEEtran}
\bibliography{paper_references}

\begin{thebibliography}{10}
\providecommand{\url}[1]{#1}
\csname url@samestyle\endcsname
\providecommand{\newblock}{\relax}
\providecommand{\bibinfo}[2]{#2}
\providecommand{\BIBentrySTDinterwordspacing}{\spaceskip=0pt\relax}
\providecommand{\BIBentryALTinterwordstretchfactor}{4}
\providecommand{\BIBentryALTinterwordspacing}{\spaceskip=\fontdimen2\font plus
\BIBentryALTinterwordstretchfactor\fontdimen3\font minus \fontdimen4\font\relax}
\providecommand{\BIBforeignlanguage}[2]{{%
\expandafter\ifx\csname l@#1\endcsname\relax
\typeout{** WARNING: IEEEtran.bst: No hyphenation pattern has been}%
\typeout{** loaded for the language `#1'. Using the pattern for}%
\typeout{** the default language instead.}%
\else
\language=\csname l@#1\endcsname
\fi
#2}}
\providecommand{\BIBdecl}{\relax}
\BIBdecl

\bibitem{feng2026tessera}
Z.~Feng \emph{et~al.}, ``{TESSERA}: Temporal embeddings of surface spectra for {E}arth representation and analysis,'' in \emph{Proceedings of the IEEE/CVF Conference on Computer Vision and Pattern Recognition (CVPR)}, June 2026, pp. 34\,818--34\,831.

\bibitem{brown2025alphaearth}
C.~F. Brown \emph{et~al.}, ``{AlphaEarth Foundations}: An embedding field model for accurate and efficient global mapping from sparse label data,'' 2025, arXiv preprint arXiv:2507.22291.

\bibitem{feng2026applications}
\BIBentryALTinterwordspacing
Z.~Feng \emph{et~al.}, ``Applications of the {TESSERA} geospatial foundation model to diverse environmental mapping tasks,'' SSRN preprint 6142416, 2026. [Online]. Available: \url{https://ssrn.com/abstract=6142416}
\BIBentrySTDinterwordspacing

\bibitem{lisaius2024barlow}
M.~C. Lisaius \emph{et~al.}, ``Using {Barlow Twins} to create representations from cloud-corrupted remote sensing time series,'' \emph{IEEE Journal of Selected Topics in Applied Earth Observations and Remote Sensing}, vol.~17, pp. 13\,162--13\,168, 2024.

\bibitem{ball2026species}
J.~G.~C. Ball \emph{et~al.}, ``Geospatial foundation models enable data-efficient tree species mapping in temperate mountain forests,'' \emph{Science of Remote Sensing}, vol.~14, p. 100466, 2026.

\bibitem{lisaius2026senegal}
M.~C. Lisaius \emph{et~al.}, ``Embedding-based crop type classification in the groundnut basin of {Senegal},'' 2026, arXiv preprint arXiv:2601.16900.

\bibitem{knezevic2026wildfire}
J.~Knezevic \emph{et~al.}, ``Annual {E}arth-observation embeddings encode wildfire disturbance and support simplified burned area mapping,'' 2026, arXiv preprint arXiv:2609.25731.

\bibitem{feng2026tesserav2}
Z.~Feng \emph{et~al.}, ``{TESSERA} v2: Scaling pixel-wise {E}arth foundation models,'' 2026, arXiv preprint arXiv:2607.03949.

\bibitem{feng2026hpc}
\BIBentryALTinterwordspacing
Z.~Feng \emph{et~al.}, ``Operational {HPC} lessons learnt training the {Tessera} geospatial foundation model series,'' University of Cambridge, Cambridge, United Kingdom, Tech. Rep., Sep. 2026. [Online]. Available: \url{https://www.cl.cam.ac.uk/~avsm2/tessera-hpc-report-20260922.pdf}
\BIBentrySTDinterwordspacing

\bibitem{richards1984pca}
J.~A. Richards, ``Thematic mapping from multitemporal image data using the principal components transformation,'' \emph{Remote Sensing of Environment}, vol.~16, no.~1, pp. 35--46, 1984.

\bibitem{fung1987pca}
T.~Fung and E.~LeDrew, ``Application of principal components analysis to change detection,'' \emph{Photogrammetric Engineering and Remote Sensing}, vol.~53, pp. 1649--1658, 1987.

\bibitem{zhu2014ccdc}
Z.~Zhu and C.~E. Woodcock, ``Continuous change detection and classification of land cover using all available {Landsat} data,'' \emph{Remote Sensing of Environment}, vol. 144, pp. 152--171, 2014.

\bibitem{chen2024ttp}
K.~Chen \emph{et~al.}, ``Time travelling pixels: Bitemporal features integration with foundation model for remote sensing image change detection,'' in \emph{IGARSS 2024 - 2024 IEEE International Geoscience and Remote Sensing Symposium}, 2024, pp. 8581--8584.

\bibitem{li2026forecasting}
J.~Li \emph{et~al.}, ``Forecasting {E}arth surface change in embedding space,'' 2026, arXiv preprint arXiv:2611.02211.

\bibitem{lcmap_reference}
{U.S. Geological Survey}, ``{LCMAP} {CONUS} reference data,'' \url{https://www.usgs.gov/special-topics/lcmap/lcmap-conus-reference-data}.

\bibitem{mtbs_overview}
{U.S. Geological Survey} and {USDA Forest Service}, ``Monitoring trends in burn severity: Project overview,'' \url{https://www.mtbs.gov/project-overview}.

\bibitem{jolliffe2016pca}
I.~T. Jolliffe and J.~Cadima, ``Principal component analysis: A review and recent developments,'' \emph{Philosophical Transactions of the Royal Society A: Mathematical, Physical and Engineering Sciences}, vol. 374, no. 2065, p. 20150202, 2016.

\bibitem{roberts2017crossvalidation}
D.~R. Roberts \emph{et~al.}, ``Cross-validation strategies for data with temporal, spatial, hierarchical, or phylogenetic structure,'' \emph{Ecography}, vol.~40, no.~8, pp. 913--929, 2017.

\bibitem{mtbs_products}
{U.S. Geological Survey} and {USDA Forest Service}, ``{MTBS} product descriptions,'' \url{https://www.mtbs.gov/product-descriptions}.

\bibitem{usgs2022lcmap13}
\BIBentryALTinterwordspacing
{U.S. Geological Survey}, ``Land change monitoring, assessment, and projection ({LCMAP}) collection 1.3 science products for the conterminous {U}nited {S}tates,'' USGS data release, 2022. [Online]. Available: \url{https://doi.org/10.5066/P9C46NG0}
\BIBentrySTDinterwordspacing

\bibitem{hastie2009esl}
T.~Hastie \emph{et~al.}, \emph{The Elements of Statistical Learning: Data Mining, Inference, and Prediction}, 2nd~ed.\hskip 1em plus 0.5em minus 0.4em\relax New York, NY, USA: Springer, 2009.

\bibitem{efron1979bootstrap}
B.~Efron, ``Bootstrap methods: Another look at the jackknife,'' \emph{The Annals of Statistics}, vol.~7, no.~1, pp. 1--26, 1979.

\bibitem{guerrero2026temporal}
J.~Guerrero-Viu \emph{et~al.}, ``Temporal sensitivity analysis of {Tessera} embeddings,'' 2026, arXiv preprint arXiv:2608.27175.

\bibitem{young2026earth}
R.~Young and S.~Keshav, ``{E}arth embeddings are decomposable records of the land surface,'' 2026, arXiv preprint arXiv:2610.32617.

\bibitem{benson2026decomposing}
V.~Benson \emph{et~al.}, ``Decomposing {E}arth embeddings with sparse autoencoders,'' 2026, arXiv preprint arXiv:2611.11398.

\end{thebibliography}

\end{document}